\documentclass[letterpaper, 10 pt, conference]{ieeeconf}

\IEEEoverridecommandlockouts
\usepackage{amsmath}
\usepackage{amssymb}
\usepackage{graphicx}
\usepackage{cuted}
\usepackage{capt-of}
\usepackage{booktabs}
\usepackage{makecell}
\usepackage{multirow}
\usepackage{threeparttable}
\usepackage{algorithm}
\usepackage{algpseudocode}
\usepackage{cite}
\usepackage{placeins}
\usepackage{balance}
\usepackage[hidelinks]{hyperref}

\title{\LARGE \bf
Learning Soccer Shooting Skills for Humanoid Robots: A Progressive Perception-Action Framework
}
\author{\authorblockN{Jipeng Kong\textsuperscript{1,2*} \quad Xinzhe Liu\textsuperscript{1,2*}
\quad Yuhang Lin\textsuperscript{1,3}
\quad Jinrui Han\textsuperscript{1,4} \quad \\
Jiyuan Shi\textsuperscript{1} \quad S\"oren Schwertfeger\textsuperscript{2} \quad Chenjia Bai\textsuperscript{1\dag{}} \quad Xuelong Li\textsuperscript{1\dag{}} 
}
\authorblockA{
\textsuperscript{1}Institute of Artificial Intelligence (TeleAI), China Telecom \quad 
\textsuperscript{2}ShanghaiTech University \quad \\
\textsuperscript{3}Zhejiang University \quad 
\textsuperscript{4}Shanghai Jiao Tong University \quad  \\
\textsuperscript{$*$}Equal contribution \quad
\textsuperscript{\dag{}}Corresponding author
}
}

\begin{document}

\maketitle
\thispagestyle{empty}
\pagestyle{empty}

\begin{strip}
  \centering
  \includegraphics[width=\linewidth]{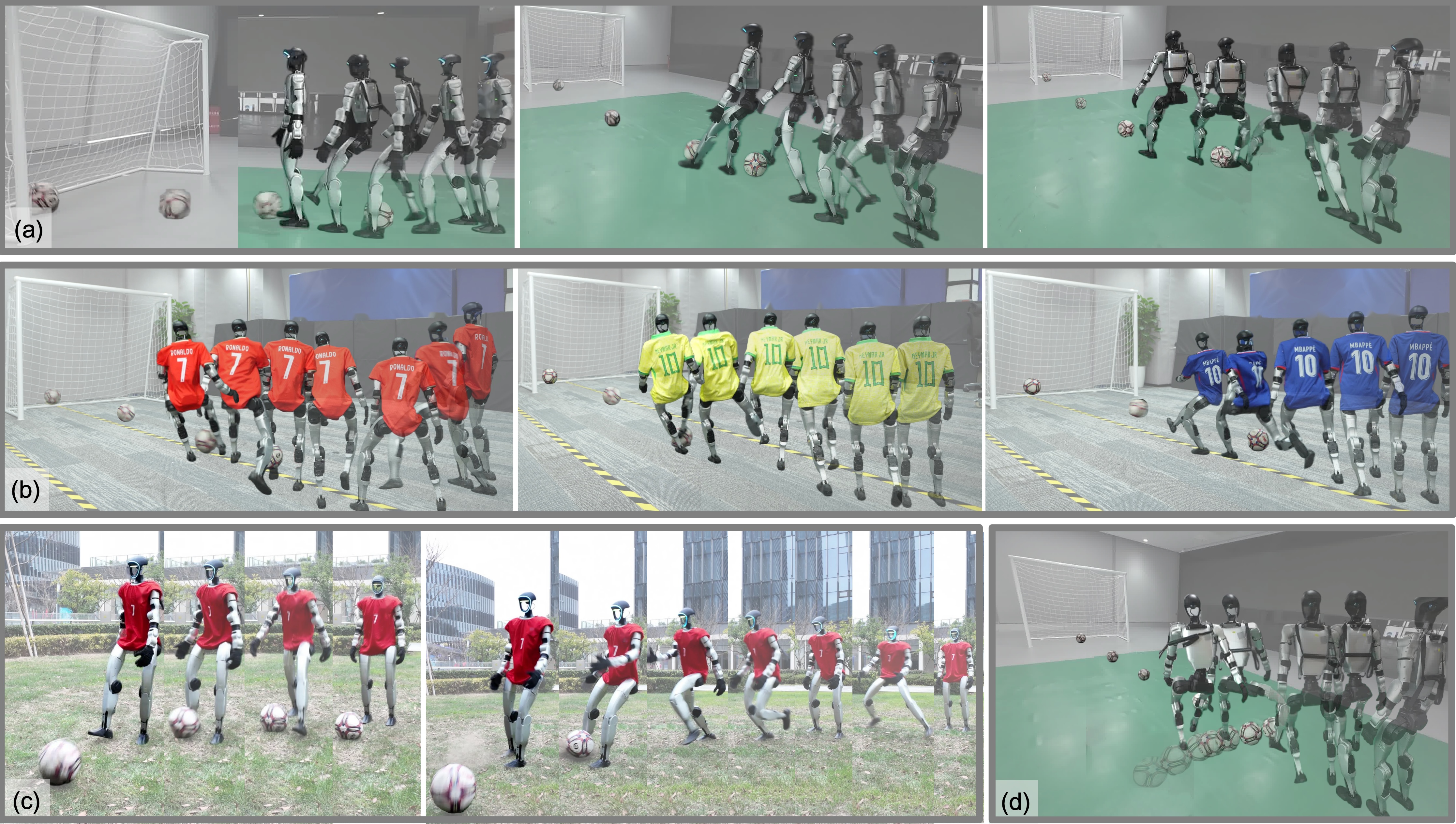}
  \captionof{figure}{Humanoid learning soccer shooting skills. (a) The robot performs goal-directed kicks referring to different motions conditioned on the ball position.
(b) The robot achieves accurate shots while imitating styles of different professional players.
(c) Stable and precise kicking on a grass field.
(d) The robot can successfully kick a moving ball.}
  \label{fig:big_real_picture}

\end{strip}

\begin{abstract}
Soccer presents a significant challenge for humanoid robots, demanding tightly integrated perception-action capabilities for tasks like perception-guided kicking and whole-body balance control. Existing approaches suffer from inter-module instability in modular pipelines or conflicting training objectives in end-to-end frameworks. We propose Perception-Action integrated Decision-making (PAiD), a progressive architecture that decomposes soccer shooting skill acquisition into three components: motion-skill acquisition via human motion tracking, lightweight perception-action integration for positional generalization, and physics-aware sim-to-real transfer. This decomposition establishes stable foundational skills, mitigates reward conflicts during perception integration, and minimizes sim-to-real gaps. Experiments on the Unitree G1 demonstrate high-fidelity human-like kicking with robust performance under diverse conditions—including static or rolling balls, various positions, and disturbances—while maintaining consistent execution across indoor and outdoor scenarios. Our divide-and-conquer strategy advances robust humanoid soccer capabilities and offers a scalable framework for complex embodied skill acquisition.
\end{abstract}


\section{Introduction}

Soccer represents a globally recognized benchmark for evaluating the integrated perception-action capabilities of humanoid robots \cite{kitano1997robocup}. Unlike the miniature Robotis OP3 platform (height: 0.5 m, mass: 3.5 kg) \cite{ScienceRobotics}, today's humanoid robots stretch well beyond 1.2 m in height and 30 kg in mass. This scale escalation intensifies challenges in whole-body balancing and human-like kicking motions due to gravitational and inertial effects \cite{liao2024berkeley,xie2025humanoid, shi2025adversarial}. Meanwhile, unlike ordinary humanoid locomotion tasks \cite{HugWBC,HumanGym1}, learning soccer shooting demands real-time adaptation to unstructured environments: robots must simultaneously process dynamic scene changes, regulate whole-body locomotion, and execute precise ball-kicking maneuvers while maintaining balance. This holistic fusion of perception and action integrates multiple core competencies, including dynamic visual perception, whole-body control, and spatiotemporally accurate striking.

Existing approaches predominantly adopt two paradigms: (i) modular hierarchical architectures that segregate perception, motion planning, and low-level control into decoupled modules \cite{ji2023dribblebot,abreu2019runfaster}; and (ii) end-to-end reinforcement learning (RL) frameworks pursuing joint optimization of visual inputs to motor control \cite{agile-striker,soccer-quad}. The former paradigm often incurs inter-module representation gaps, while the latter exhibits training instability stemming from reward conflicts between locomotion, ball kicking, and post-impact recovery. Critically, although recent vision-driven successes \cite{soccer-learning} have exploited adversarial motion priors (AMP) \cite{peng2021amp}, they still depend on manually engineered rewards, severely limiting scalability and real-world robustness.

To address these challenges, we propose \emph{Perception-Action integrated Decision-making (PAiD)}, a progressive architecture for learning humanoid soccer shooting skills. Our core insight is to rethink skill acquisition: complex behaviors should not be treated as monolithic optimization problems but rather acquired through a curriculum that first masters foundational motions, then integrates perceptual feedback, and finally solves real-world deployment. Specifically: (i) At the \emph{motion-skill acquisition} stage, we adopt whole-body motion tracking to learn various soccer shooting skills in diverse styles from human demonstrations. The robot learns natural, high-fidelity soccer shooting behaviors for various goal locations. Crucially, this stage decouples how to kick from where to kick by isolating low-level policy learning from positional randomization, establishing a stable reference-aligned motion prior. (ii) Then, we perform \emph{perception-action integration} by introducing lightweight environmental perception for reaching and kicking balls in different positions. In this stage, the robot learns to chase and kick positionally generalizable targets and even rolling balls. A minimal prior reward guides the policy to adjust whole-body gaits and orientation toward the ball. Due to the distinct stage separation with lightweight perception modules, our method mitigates reward conflicts inherent in single-stage methods. (iii) For \emph{sim-to-real transfer}, we identify that minor discrepancies in physics properties (e.g., restitution and friction of the ball) between simulation and the real world drastically degrade performance. We thus propose an iterative physics alignment strategy to identify key parameters of the ball in simulation to ensure efficient sim-to-real alignment. For real-world perception, we combine visual and lidar-based localization to obtain perception input.

We validate PAiD on a Unitree G1 humanoid platform. The robot exhibits robust kicking capabilities, achieving a high shooting success rate under diverse ball positions, lighting variations, and physical disturbances. Our method achieves superior sim-to-real transfer, where the physics alignment strategy yields better sample efficiency than baselines, with consistent performance across indoor and outdoor surfaces.

Our contributions are threefold: (i) A novel learning framework that decomposes soccer shooting skills into two progressive stages of motion acquisition and perception-action integration, with sim-to-real transfer incorporated into the second training stage; (ii) A versatile motion tracking system for learning diverse human kicking motions and a lightweight perception integration mechanism ensuring accurate ball interaction; (iii) An alignment strategy that resolves physical and visual gaps between simulation and real-world deployment. Our work advances humanoid soccer capabilities and provides a divide-and-conquer strategy for perception-based human-skill acquisition in embodied agents.

\begin{figure*}[!t]
    \centering

    \includegraphics[width=0.90\linewidth]{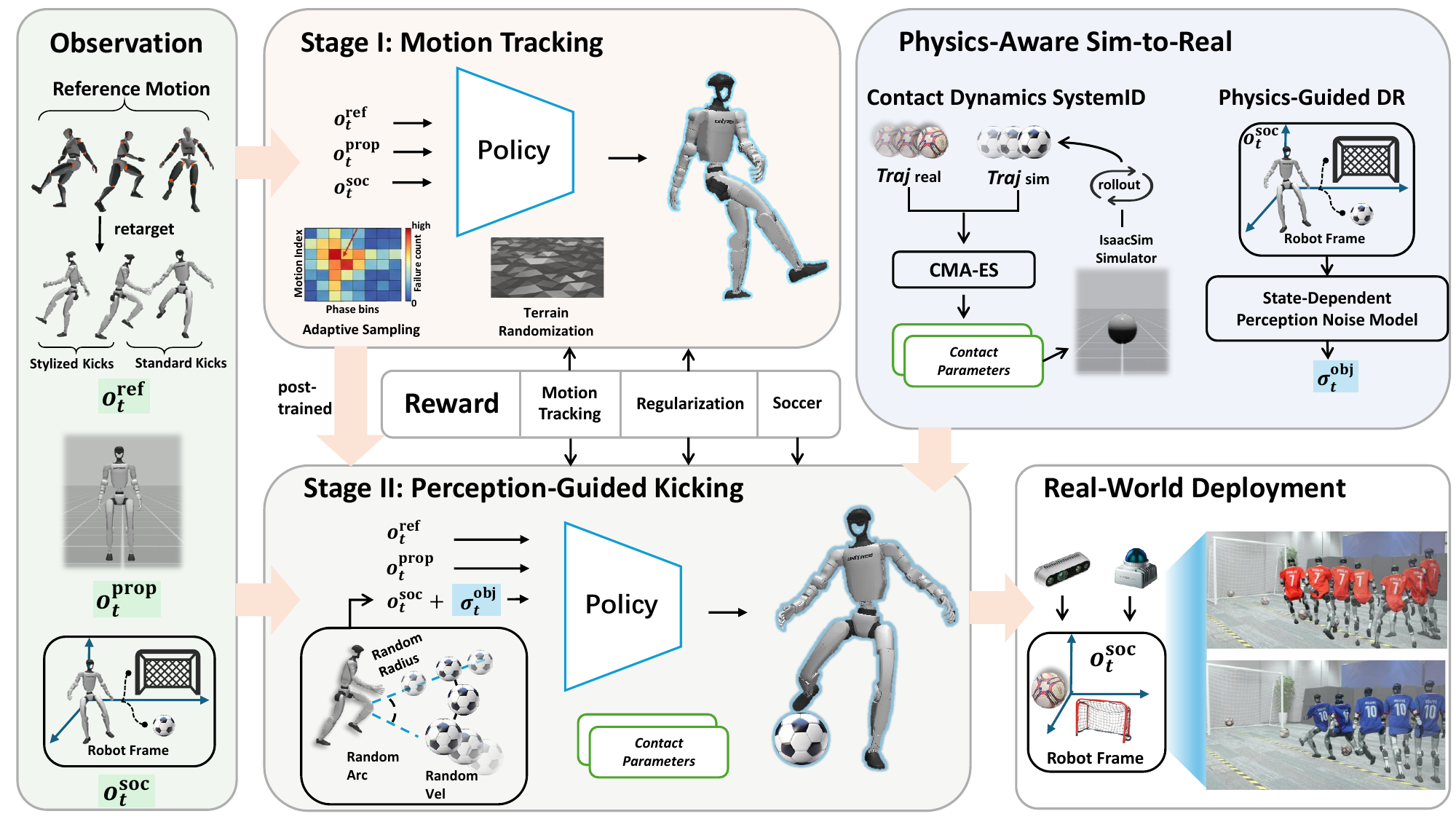}
    \caption{Overview of the Perception-Action integrated Decision-making (PAiD) framework. Our pipeline progressively acquires robust soccer shooting skills through three components: (1) \textbf{Motion Tracking}: We retarget diverse human kicking motions (Standard \& Stylized) to the humanoid and train a unified tracking policy using adaptive sampling to master fundamental skills under reference-aligned setups. (2) \textbf{Perception-Guided Kicking}: We equip the policy with egocentric perception and task-specific rewards to generalize kicking skills to randomized static and rolling ball targets. (3) \textbf{Physics-Aware Sim-to-Real Transfer}: We bridge the reality gap by aligning simulation contact dynamics with real-world measurements (ball drop \& rolling tests) and incorporating physics-guided observation noise. (4) \textbf{Real-World Deployment} We successfully deploy PAiD on the Unitree G1.}
    \label{fig:framework_overview}

\end{figure*}

\section{Related Works}

\paragraph{Humanoid Whole-Body Control} Proprioception-driven reinforcement learning enables diverse locomotion and whole-body behaviors~\cite{HugWBC,he2025learning,long2024learning}, while adversarial motion priors encourage human-like styles~\cite{peng2021amp,MobileTele}. Motion-tracking approaches instead learn from retargeted human motions~\cite{GMR}, using reference-based rewards to reproduce whole-body skills~\cite{deepmimic,asap,ExBody,kungfubot,liao2025beyondmimic}. General motion-tracking systems further extend imitation to whole-body teleoperation~\cite{ze2025twist2,han2025kungfubot2,GMT,zhang2025track}. However, reference tracking alone does not address perception-guided adaptation to changing ball positions. Perception-integrated approaches explore terrain awareness and human--object interaction~\cite{intermimic,weng2025hdmi,ResMimic,yang2025omniretarget}, including reconstruction-based control in VideoMimic~\cite{videoMimic} and hierarchical planning and control in HITTER~\cite{HITTER}. Our focus is on extending human motion tracking to soccer, where the policy must adapt its approach and strike to the observed ball position while retaining the reference motion style. We therefore separate motion-skill acquisition from perception-guided positional generalization.

\paragraph{Robot Soccer} Quadruped soccer studies combine motion priors, planning, and reinforcement learning for shooting and goalkeeping~\cite{Huang2022CreatingAD,Ji2022HierarchicalRL,soccer-quad}, but do not directly address human-like kicking on humanoid robots. Humanoid approaches investigate chasing and kicking~\cite{agile-striker}, locomotion and dribbling~\cite{ScienceRobotics}, active perception~\cite{Wang2025DribbleML}, and environment reconstruction~\cite{Tirumala2024LearningRS}. These studies motivate integrating ball perception with coordinated whole-body control. Closely related work~\cite{soccer-learning} combines adversarial motion priors with perception for kicking, optimizing perception and control in a single stage. In contrast, PAiD first acquires reference-aligned kicking skills through explicit motion tracking, then introduces randomized ball positions and lightweight task rewards. This progressive design aims to mitigate conflicts between motion imitation and task objectives while preserving diverse human kicking styles and adapting to static or rolling balls.

\section{Method}

Our goal is to enable a humanoid robot to execute precise, human-like kicks toward a target from arbitrary ball positions. To mitigate reward conflicts between objectives, we adopt a two-stage training strategy. In Stage~I, the robot learns diverse human-like soccer shooting skills via a unified motion tracking framework. In Stage~II, we generalize the policy through randomized ball placements along with a few task rewards, thus enabling the robot to kick the ball from arbitrary locations to the target, while maintaining human motion styles. Notably, our framework achieves this using only a compact dataset of reference motions. Finally, we perform physics-aware system identification and domain randomization to facilitate sim-to-real transfer.

The policy observation is $\mathbf{o}_t=[\mathbf{o}^{\text{prop}}_t,\mathbf{o}^{\text{ref}}_t,\mathbf{o}^{\text{soc}}_t]$, combining proprioception, reference-motion targets, and a compact soccer context. The soccer context $\mathbf{o}^{\text{soc}}_t$ consists of the ball position $\mathbf{g}_{ball}=\mathbf{R}_{\text{pelvis}}^{-1}(\mathbf{p}^{w}_{\text{ball}}-\mathbf{p}^{w}_{\text{pelvis}})$ and goal position $\mathbf{g}_{goal}=\mathbf{R}_{\text{pelvis}}^{-1}(\mathbf{p}^{w}_{\text{goal}}-\mathbf{p}^{w}_{\text{pelvis}})$ relative to the robot's root frame.
During Stage~I, the ball is placed according to the reference motion and the goal is set at the nominal target location, providing a reference-aligned setup for learning stable kicking primitives. In Stage~II, the ball placement is randomized to enable perception-guided positional generalization. In real-world deployment, the ball position is obtained via fused visual and lidar-based localization, so $\mathbf{g}_{ball}$ is continuously available.

\subsection{Stage~I: Soccer-skill Acquisition via Motion Tracking}

The primary objective of this stage is to acquire high-fidelity, physically feasible kicking primitives that are robust across diverse terrains. By learning these skills in a reference-aligned setup, the policy can focus on reproducing structured kicking behaviors, which later serve as a stable motion prior for perception-guided generalization in Stage~II.

\paragraph{Data Preparation}
To ensure the policy masters a diverse repertoire of kicking skills and covers a broad soccer shooting zone, we curated a specialized motion dataset comprising 13 distinct human soccer kicking motions using a motion capture system. The dataset is structured into two categories: (i) \textit{standard kicks}, which are recorded with uniformly distributed ball placements (approximately 30 cm intervals) to cover diverse distances and shooting angles; and (ii) \textit{stylized kicks}, which mimic signature moves of famous soccer players. For each motion, the striking leg was naturally selected based on the ball's relative location to ensure biomechanical feasibility. These motions are then retargeted to the robot via GMR~\cite{GMR} and annotated with the \textit{striking leg} (left or right). This information is vital for the subsequent stage, serving as a gating signal for the contact-based rewards, ensuring the robot learns to strike the ball with the intended foot.
The retargeted motions cover approximately $x\in[-0.5,1.3]$\,m and $y\in[0,1.5]$\,m.

\paragraph{Unified Motion Tracking}
We follow BeyondMimic~\cite{liao2025beyondmimic} to compute a yaw-only alignment $\Delta\mathbf{R}_t$ from the relative anchor rotation and use it to rotate the reference motion into the robot's current anchor yaw before evaluating tracking errors. This alignment facilitates the robot in adjusting its heading to strike the ball towards various angles.
However, relying on a single reference motion causes overfitting and restricts positional generalization.

To tackle this challenge, we extend the BeyondMimic formulation to a multi-motion soccer-kicking setting by adding the motion index to the adaptive sampling space. Specifically, we maintain a failure histogram $F \in \mathbb{R}^{M \times B}$ over motion indices and phase bins, and sample motion-phase pairs according to the smoothed failure distribution. This steers training toward difficult segments across both standard and stylized kicks. Meanwhile, we incorporate terrain randomization -- including irregular and uneven surfaces -- directly into Stage~I training. This not only enables the policy to adapt to a wide range of terrains but also removes the need to explicitly handle terrain-related difficulty when designing rewards in Stage~II.

Regarding reward design for the motion-tracking stage and the subsequent perception-guided stage, we employ motion-tracking rewards along with regularization in Stage~I. We set the episodes to terminate either upon reaching a time limit or when tracking violates safety constraints, e.g., due to large attitude errors inferred from projected gravity, substantial deviations in anchor height, or pronounced kinematic inconsistencies. As a result, Stage~I yields stable, human-like kicking primitives that can later be specialized to different ball setups with only minimal additional objectives.

\subsection{Stage~II: Perception-Guided Positional Generalization}

The primary objective of this stage is to learn perception-guided kicking behaviors that generalize across varying ball positions and reliably produce goal-directed shots.
To this end, we resume the Stage~I policy, retain motion tracking as a structural prior and introduce a lightweight perception integration mechanism, while mitigating interference with the learned kicking dynamics.

\paragraph{Ball Position Sampling}
For each selected motion, we first place the ball at a motion-consistent nominal location determined by its terminal position, and then sample randomized ball spawns around this nominal location by (i) applying an angular perturbation around the nominal direction and (ii) sampling a radius offset. This yields a curved arc of feasible ball locations around the intended kick direction, promoting positional generalization without disrupting the motion timing. 
Notably, we sample an initial linear velocity for the ball within a small range, and we employ an LSTM-based policy architecture capable of temporal aggregation. Consequently, the policy can implicitly predict the short-horizon trajectory of rolling balls, ensuring consistent striking performance despite ball motion.

\paragraph{Lightweight Perception-Action Rewards}
We retain motion tracking but disable anchor-position tracking to allow positional adaptation. Five task rewards guide the approach, strike, and ball motion, as summarized in Table~\ref{tab:lightweight_rewards}.

Before contact, ball proximity rewards a small horizontal pelvis-to-ball distance $d_t$ at time step $t$. This reward is frozen at the first valid contact time $t_c$, with $t_c=+\infty$ if no valid contact occurs. A valid contact is the first ball contact made by the motion-labeled striking foot; $c_t$ is one at that step and zero otherwise. At contact, the strike-direction reward favors the leg-specific lateral swing direction $s_{\mathrm{leg}}\in\{-1,+1\}$. Here, $\widehat v^{\mathrm{foot}}_{y,t}$ denotes the lateral component of the normalized striking-foot velocity in the robot frame.

After a valid strike, velocity alignment and ball speed encourage accurate, forceful shots. The angular error $\theta_t$ measures the deviation of the planar ball velocity from the initial ball-to-goal direction, and $u_t$ denotes planar ball speed. The window weights $w_t^{\mathrm{align}}$ and $w_t^{\mathrm{speed}}$ activate these rewards only within the post-contact reward window and when the minimum ball-speed condition is met.


\begin{table}[!htbp]
    \centering
    \caption{Lightweight perception-action rewards in Stage~II.}
    \label{tab:lightweight_rewards}
    \begin{tabular}{lcc}
        \toprule
        Term & \makecell{Expression} & Weight \\
        \midrule
        \makecell[l]{Ball approach} & $\displaystyle
        \begin{cases}
            \exp\!\left(-\frac{d_t^2}{4^2}\right), & t<t_c,\\
            \exp\!\left(-\frac{d_{t_c}^2}{4^2}\right), & t\geq t_c
        \end{cases}$ & 1.0 \\
        \addlinespace
        Desired Contact & $\displaystyle c_t=\begin{cases}
            1, & \text{valid first contact},\\
            0, & \text{otherwise}
        \end{cases}$ & 50.0 \\
        \addlinespace
        \makecell[l]{Strike direction} & $c_t\max\!\left(s_{\mathrm{leg}}\widehat v^{\mathrm{foot}}_{y,t},0\right)$ & 50.0 \\
        \addlinespace
        \makecell[l]{Velocity alignment} & $\displaystyle w_t^{\mathrm{align}}\exp\!\left(-\frac{\theta_t^2}{0.8^2}\right)$ & 30.0 \\
        \addlinespace
        \makecell[l]{Ball speed} & $\displaystyle w_t^{\mathrm{speed}}\left[1-\exp\!\left(-\frac{u_t^2}{1.2^2}\right)\right]$ & 10.0 \\
        \bottomrule
    \end{tabular}
\end{table}

\paragraph{Stabilization Terms}
To maintain balance under perception-driven adjustments, we include stabilization terms such as waist action-rate regularization, pelvis uprightness (penalizing roll/pitch deviation via projected gravity), and a foot-separation reward to reduce foot crossing. These additions are deliberately lightweight, helping preserve the learned human-like kick kinematics while mitigating the unstable coupled optimization commonly observed in perception-to-control training.

Overall, the synergy of these mechanisms and rewards enables the robot to achieve robust and precise human-like kicking capabilities while ensuring effective positional generalization with only a compact set of reference motions.

\subsection{Physics-Aware Sim-to-Real Transfer}

Accurate ball dynamics are critical for real-world transfer, since minor discrepancies in contact properties such as friction or restitution can lead to large deviations in the ball trajectory. In addition, ball and goal observations are inherently noisy during deployment. To this end, we adopt a physics-aware sim-to-real strategy that combines contact dynamics identification with structured domain randomization (DR), injecting physical priors at both the dynamics and observation levels.

\paragraph{Contact Dynamics Identification}
To identify ball--ground contact dynamics, we conduct real-world drop and rolling experiments to characterize normal impact behavior and tangential frictional energy loss, respectively. The same experiments are reproduced in IsaacSim~\cite{NVIDIA_Isaac_Sim} under identical initial conditions, and we optimize a compact set of simulator parameters, including static and dynamic friction, restitution, and linear and angular damping. Let
$\mathbf{h}=\{h_i\}_{i=0}^{N_d}$ and $\mathbf{d}=\{d_i\}_{i=0}^{N_r}$ denote the measured height and horizontal displacement trajectories, and $\mathbf{h}'=\{h'_i\}_{i=0}^{N_d}$ and $\mathbf{d}'=\{d'_i\}_{i=0}^{N_r}$ denote their simulated counterparts. The parameters are identified by minimizing
\begin{equation}
\mathcal{L}_{\mathrm{sysid}}
=
\lambda_1 \sum_{i=0}^{N_d} (h_i-h_i')^2
+
\lambda_2 \sum_{i=0}^{N_r} (d_i-d_i')^2 ,
\label{eq:sysid_loss}
\end{equation}
where $\lambda_1$ and $\lambda_2$ balance impact and rolling dynamics. We solve this derivative-free optimization problem using CMA-ES~\cite{hansen2001cma}.

\begin{figure}[!htbp]  \centering \includegraphics[width=\linewidth]{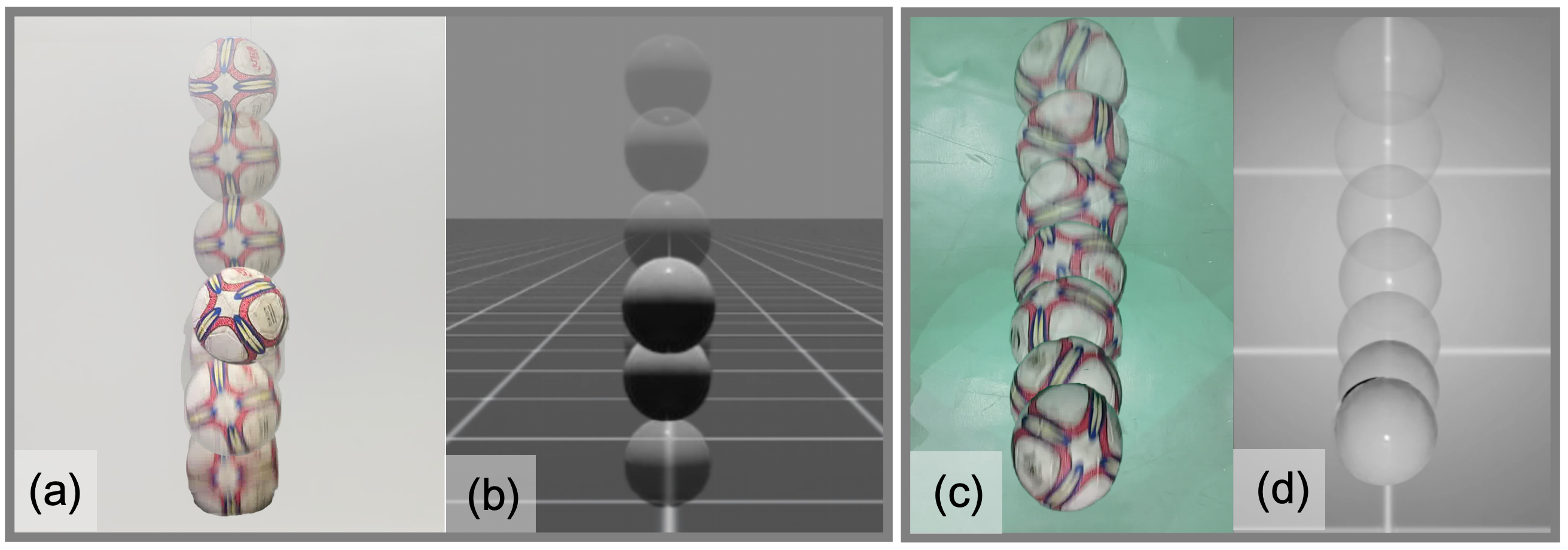}  \caption{System identification of soccer-ball dynamics. (a)--(b) ball drops; (c)--(d) rolling tests.} \label{fig:sysID}  \end{figure}

System identification is performed separately on a rigid hard ground and a soccer field surface, producing two nominal contact parameter sets that capture distinct ground conditions. As shown in Fig.~\ref{fig:sysID}, the identified parameters yield simulated ball motions that closely match real-world observations. During policy training, environments are evenly split between the two parameter sets, and each parameter is further randomized around its nominal value as
$\theta \sim \mathcal{N}(\theta_{\mathrm{nominal}}, \mathbf{I})$
to account for residual modeling errors. 






\begin{table}[!htbp]
    \centering
    \caption{Domain randomization ranges.}
    \label{tab:DR}
    \begin{tabular}{ll}
            \toprule
            Term & Range \\
            \midrule
            Robot static friction & $\mathcal{U}(0.3, 1.6)$ \\
            Robot dynamic friction & $\mathcal{U}(0.3, 1.2)$ \\
            Robot restitution & $\mathcal{U}(0.0, 0.5)$ \\
            Joint default pos & $\mathcal{U}(-0.01, 0.01)$ \\
            Base CoM & \makecell[l]{$x \sim \mathcal{U}(-0.025, 0.025)$\\$y,z \sim \mathcal{U}(-0.05, 0.05)$} \\
            Push robot & $\mathcal{U}(-0.5, 0.5)$ \\
            \bottomrule
    \end{tabular}
\end{table}

\paragraph{Physics-Guided Domain Randomization}
To account for perception uncertainty, we incorporate DR into the observation space alongside dynamics randomization.
Beyond the standard DR commonly used in motion tracking approaches \cite{han2025kungfubot2,liao2025beyondmimic}, 
we introduce a physics-guided noise model tailored to soccer ball and goal perception \cite{li2025clone,agile-striker}.

In Stage~II, the relative ball and goal positions used as observations are subject to practical estimation errors, which empirically grow with the observed object's speed and distance from the robot.
To capture this effect, we inject zero-mean Gaussian noise into object-related observations, including both the ball and the goal. For each object \( obj \in \{\text{ball}, \text{goal}\} \), the noise magnitude is defined as
\begin{equation}
\sigma^{obj}_{t} =
c_{\min} +
\frac{\lVert \mathbf{v}^{obj}_{t} \rVert}{c_{\text{vel}}} +
\frac{\lVert \mathbf{p}^{obj}_{t} - \mathbf{p}^{\text{robot}}_{t} \rVert}{c_{\text{dist}}},
\end{equation}
where \( \mathbf{p}^{obj}_{t} \) and \( \mathbf{v}^{obj}_{t} \) denote the position and velocity of object \( obj \) expressed in the robot frame. The noisy observation is then obtained by adding \( \sigma^{obj}_{t} \cdot \mathcal{N}(0, I) \).

This physics-aware sim-to-real transfer strategy exposes the policy to realistic dynamics and perception variations, improving real-world robustness.

\FloatBarrier

\begin{figure*}[!t]
    \centering
    \includegraphics[width=0.8\linewidth]{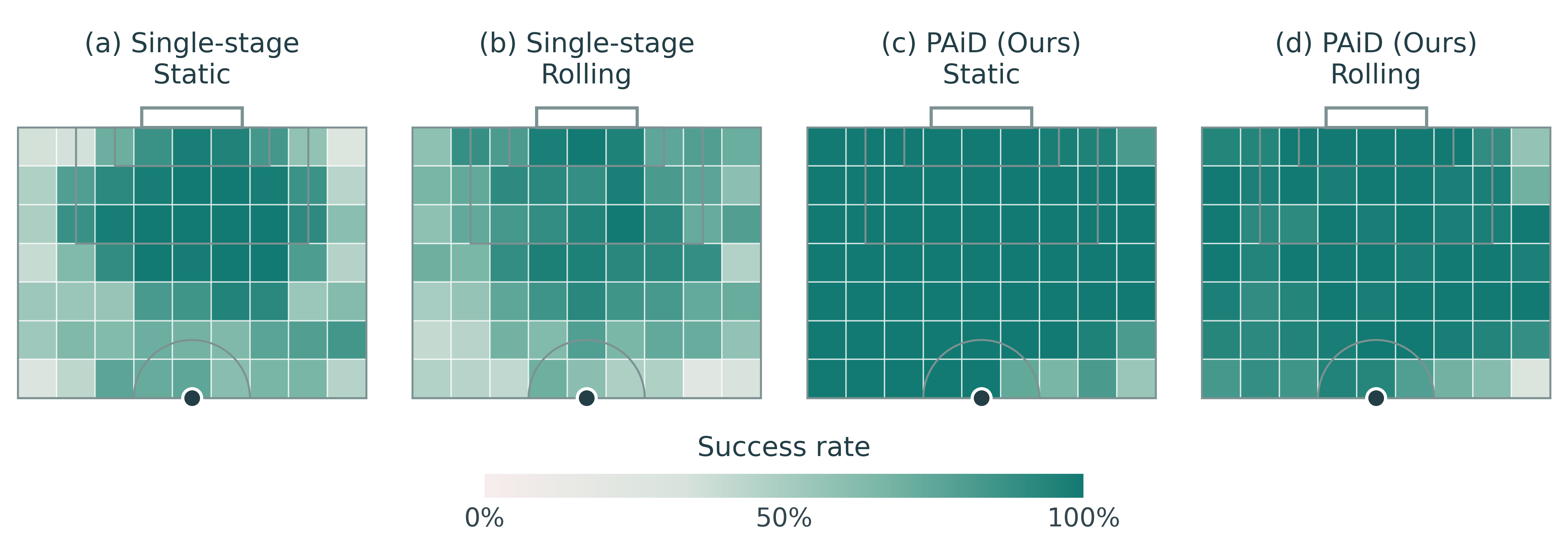}
    \caption{Half-field shooting success rates by initial ball position. (a)--(b) Single-stage and (c)--(d) PAiD, each under static-ball and rolling-ball settings, respectively.}
    \label{fig:shooting_proficiency}
\end{figure*}

\section{Experiments}

In this section, we systematically evaluate the proposed PAiD framework from three perspectives: the role of the Stage~I motion prior, soccer shooting proficiency, and real-world deployment performance. The experiments are designed to answer the following research questions. \textbf{Q1}: How do PAiD and its key components affect shooting performance and motion humanlikeness? \textbf{Q2}: How accurate and robust is the robot's soccer shooting capability? \textbf{Q3}: How well does the learned policy transfer to the physical world? For each evaluation, we specify the experimental setup and quantitative metrics in advance.

\label{subsec:motion_tracking}


\begin{table}[!htbp]
    \centering
    \caption{Shooting performance and AI-rated humanlikeness under different training schemes.}
    \label{tab:ablation_results}
    \begin{tabular}{lcccc}
        \toprule
        \textbf{Method}
        & \makecell{$E_{dis}^{ball}$(m$\downarrow$)}
        & \makecell{$E_{dir}^{ball}$(rad$\downarrow$)}
        & \makecell{$E_{succ}$ (\%$\uparrow$)}
        & \makecell{$S_H$} \\
        \midrule
        Pure RL & 1.2770 & 0.7813 & 24.92 & 2.0 \\
        AMP-based & 0.8594 & 0.5099 & 54.36 & 4.0 \\
        Single Stage & 0.6848 & 0.2877 & 71.98 & 6.0 \\
        BeyondMimic$^{*}$ & 0.6758 & 0.2101 & 76.06 & 7.5 \\
        w/o AS & 0.7272 & 0.2702 & 60.94 & 7.0 \\
        w/o LPAR
                     & 1.6716 & 0.7989 & 3.96 & 8.5 \\
        w/o ST
                     & 0.7495 & 0.2433 & 65.52 & 6.5 \\
        PAiD (Ours) & \textbf{0.2626} & \textbf{0.0757} & \textbf{99.66} & 8.0 \\
        \bottomrule
    \end{tabular}
\end{table}

\subsection{Comparative Evaluation and Ablation Study}
To answer \textbf{Q1}, we compare shooting performance and motion humanlikeness across training schemes and component ablations. We compare five training schemes: (i) Pure RL, (ii) AMP-based, (iii) Single-Stage baseline without a pretrained motion prior, (iv) BeyondMimic$^{*}$, and (v) PAiD (Ours) Stage I + Stage II. BeyondMimic$^{*}$ denotes BeyondMimic motion-tracking pretraining followed by our Stage~II soccer fine-tuning. To isolate the effect of the prior, the latter three methods are evaluated under the same Stage~II soccer setting after training. We additionally evaluate three ablations: w/o AS, w/o LPAR, and w/o ST, where AS denotes adaptive sampling, LPAR denotes lightweight perception-action rewards, and ST denotes stabilization terms; w/o indicates removal of the corresponding component. All methods are trained for the same total number of iterations.
After training, we conduct a static-ball evaluation in MuJoCo~\cite{todorov2012mujoco}, where the robot initialization is randomized uniformly within a 0.25~m radius circle, a goal with a width of 2~m is placed 5~m in front of the initialized robot, and the ball is randomly spawned within the ball-placement range $[0.5, 3.0] \times [-1.5, 1.5]$\,m.
For each evaluated policy, we run 1,000 trials per seed across five random seeds, yielding 5,000 trials per policy with identical initial states between methods.

Table~\ref{tab:ablation_results} reports four metrics. $E_{dis}^{ball}$ (m) is the mean absolute lateral deviation from the goal center at the ball's first crossing of the goal plane. $E_{dir}^{ball}$ (rad) is the mean angular deviation between the outgoing ball velocity and the direction from the impact position to the goal center. Both errors are averaged over trials with a ball strike. $E_{succ}$ (\%) is the percentage of trials ending in a goal. $S_H$ is the overall subjective AI-rated humanlikeness score on a 0--10 scale, with higher values indicating more human-like motion. In the table, $\downarrow$ and $\uparrow$ indicate that lower and higher values are better, respectively.

Table~\ref{tab:ablation_results} shows that PAiD achieves more accurate and reliable shooting than single-stage training or fine-tuning from a general motion-tracking prior, supporting the progressive acquisition of soccer-specific skills. Although the reference motion library covers robot--ball distances only up to 1.3~m, PAiD achieves an overall shooting success rate of 99.66\% at evaluation distances up to 3.0~m, demonstrating generalization beyond the reference coverage. The ablations further show that adaptive sampling and stabilization terms contribute to accurate and reliable skill execution.

We sample 20 trajectories per policy from the static-ball evaluation described above, using the same initial states across all eight policies. GPT-6 Astra evaluates sampled frame sequences rendered with matched camera settings under an identical prompt and scoring rubric, assigning each policy an overall score $S_H$ from 0 (no resemblance to human motion) to 10 (fully matching human motion structure and timing). Table~\ref{tab:ablation_results} shows that PAiD combines higher humanlikeness than Single-stage and BeyondMimic$^{*}$ with better shooting performance. Although w/o LPAR receives the highest humanlikeness score, its poor shooting performance highlights the distinction between human-like motion and effective ball interaction.

\begin{figure}[!htbp]
    \centering
    \includegraphics[width=\linewidth]{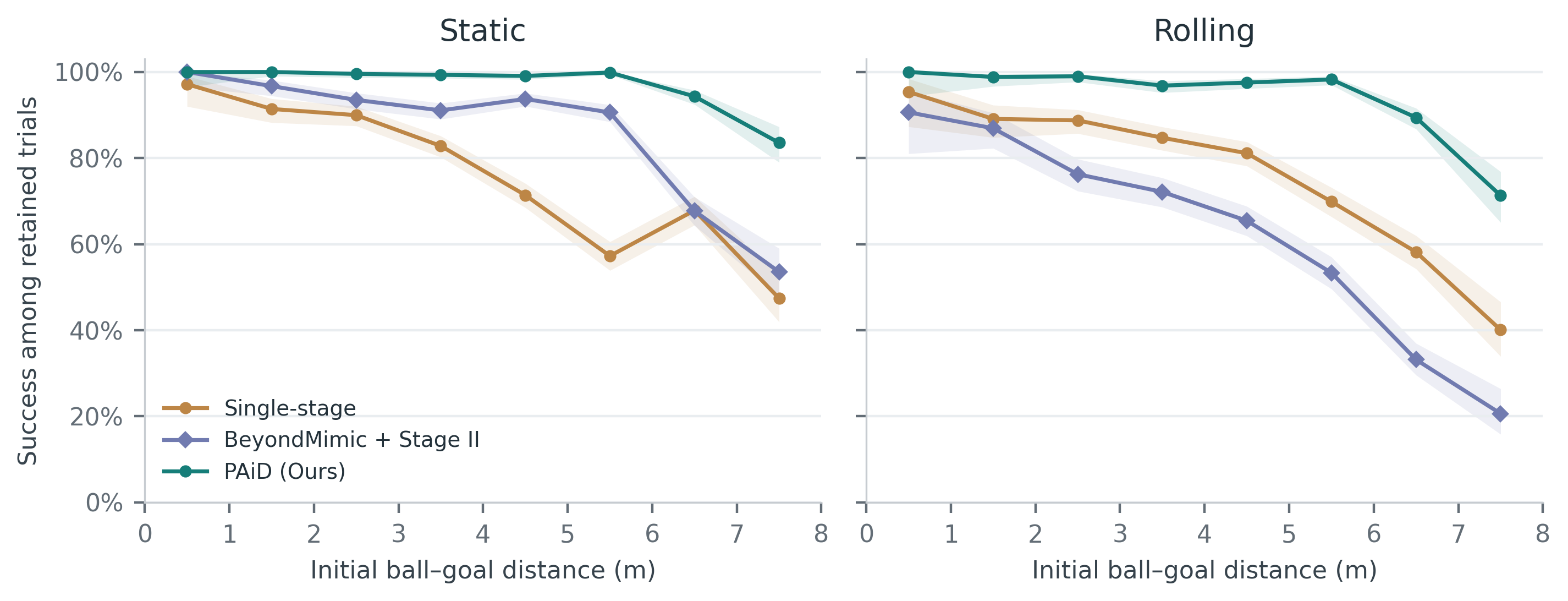}
    \caption{Shooting success versus initial distance from the ball to the target goal center for static (left) and rolling (right) balls.}
    \label{fig:checkpoint_distance_comparison}
\end{figure}

\begin{figure*}[!t]
    \centering
    \setlength{\abovecaptionskip}{3pt}
    \begin{minipage}{0.24\linewidth}
        \centering
        \includegraphics[width=\linewidth]{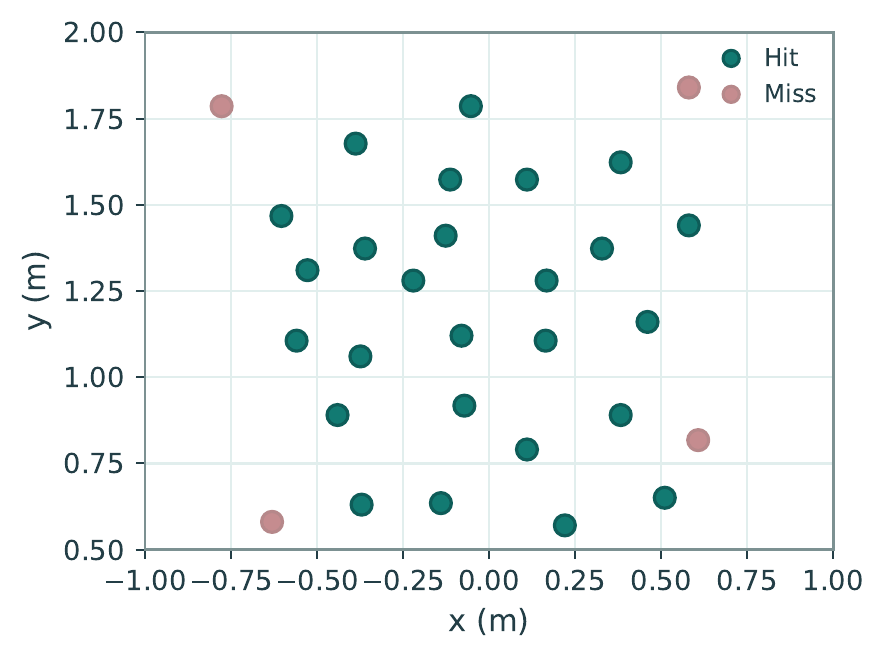}
        \centerline{\scriptsize (a) Hard ground}
    \end{minipage}\hfill
    \begin{minipage}{0.24\linewidth}
        \centering
        \includegraphics[width=\linewidth]{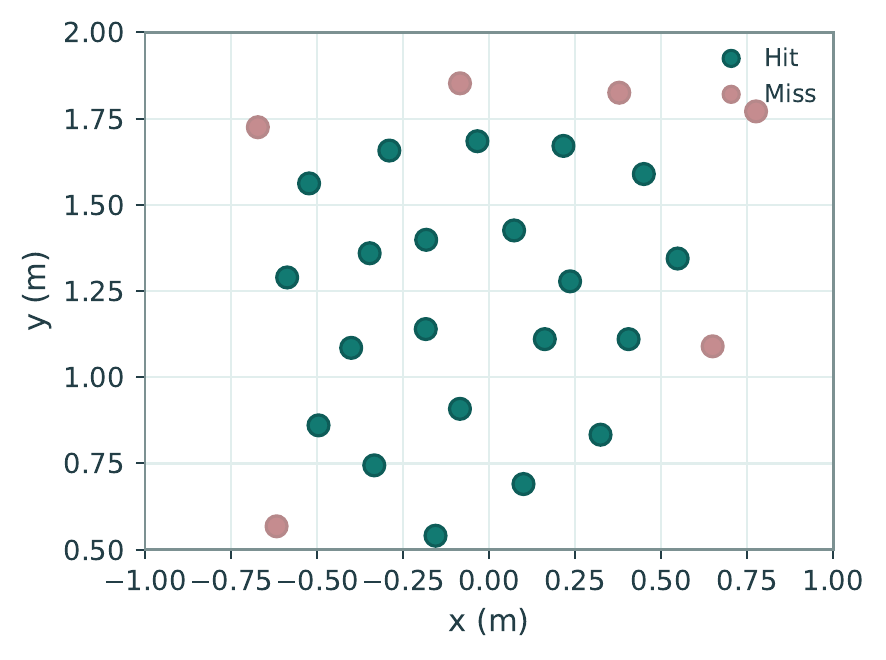}
        \centerline{\scriptsize (b) Grass ground}
    \end{minipage}\hfill
    \begin{minipage}{0.24\linewidth}
        \centering
        \includegraphics[width=\linewidth]{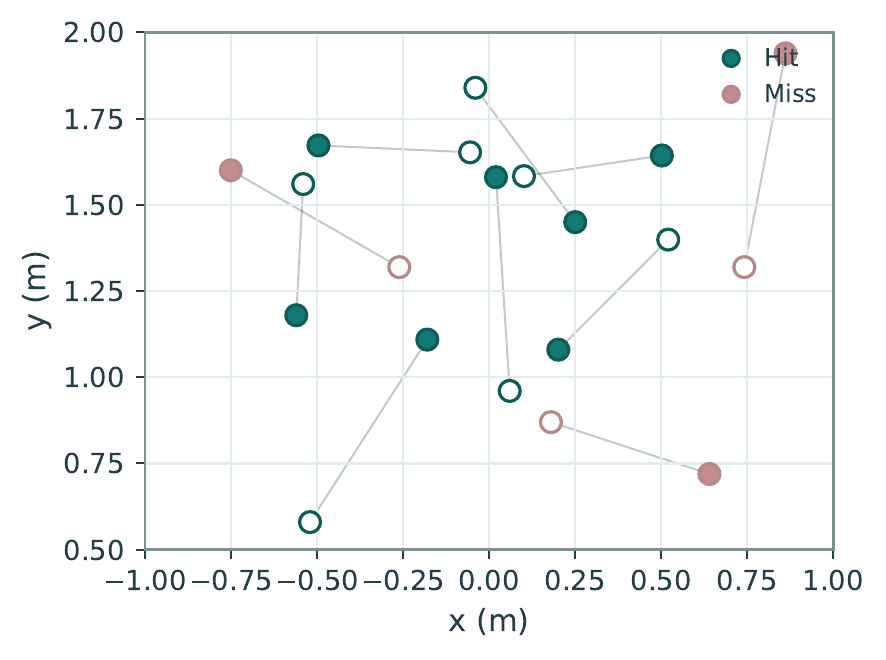}
        \centerline{\scriptsize (c) Hard ground (Rolling)}
    \end{minipage}\hfill
    \begin{minipage}{0.24\linewidth}
        \centering
        \includegraphics[width=\linewidth]{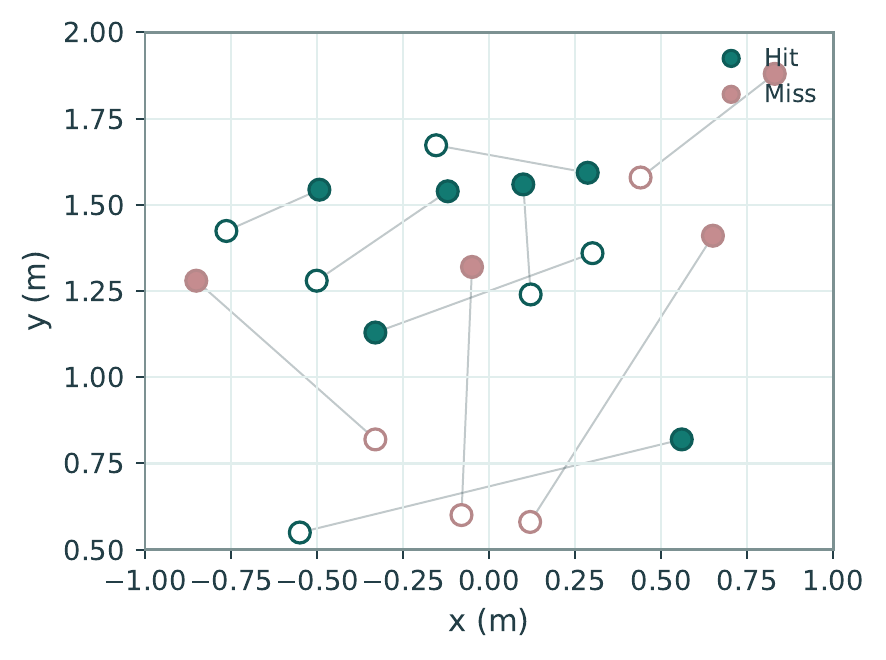}
        \centerline{\scriptsize (d) Grass ground (Rolling)}
    \end{minipage}

    \caption{Real-world shooting tests with randomly placed soccer balls. (a)–(b) show results on hard ground and grass, with 30 trials per surface. (c)–(d) show rolling-ball tests with 10 trials per surface. Hollow and solid markers denote the initial ball positions and the positions at first robot--ball contact. Teal and light pink markers indicate success and failure, respectively.}
    \label{fig:real_exp_1}
    \vspace{-6pt}
\end{figure*}

\subsection{Spatial and Dynamic Generalization}
\label{subsec:soccer_shooting}
To answer \textbf{Q2}, we compare Single-stage, BeyondMimic$^{*}$, and PAiD in MuJoCo~\cite{todorov2012mujoco}. The robot starts at the field center with a fixed heading, while ball positions are sampled uniformly across either half, targeting the goal in the ball's initial half. We evaluate static balls and rolling balls with randomized initial velocities. Placements exclude initial robot collisions but are not restricted by motion-library coverage, and all methods use identical initial states across repeated trials.

Each shooting policy is combined with the same locomotion controller, which approaches the ball and aligns the robot for shooting to evaluate shooting-policy performance over a wider range of ball positions. A phase-based handoff then starts the kick at a fixed pre-contact phase of the reference motion without resetting the robot state. All methods share the same approach and aiming rules, and each trial permits a single shooting attempt.

We report \textit{Success Rate}, defined as the fraction of evaluated trials in which the whole ball crosses the target goal line between the posts and below the crossbar after a valid kick, with no robot fall during the trial. Trials in which the ball leaves the field before contact are excluded from the success-rate calculation.

\begin{table}[!htbp]
\centering
\caption{Ablation of physics-aware sim-to-real transfer}
\label{tab:DR_exp}
\begin{threeparttable}
\begin{tabular}{lccc}
\toprule
Method & \makecell{Hard\\Ground} & \makecell{Grass\\Ground} & \makecell{$E_{succ}$ (\%)} \\
\midrule
PAiD & \textbf{39/45} & \textbf{36/45} & \textbf{83.3} \\
w/o SystemID & 33/45 & 18/45 & 56.7 \\
w/o Obs Noise & 23/45 & 20/45 & 47.8 \\
\bottomrule
\end{tabular}
\end{threeparttable}
\end{table}

Fig.~\ref{fig:shooting_proficiency} compares Single-stage and PAiD over the forward half-field, showing broader regions of high success for PAiD. Fig.~\ref{fig:checkpoint_distance_comparison} further compares all three methods across initial ball--goal distances using trials from both halves. BeyondMimic$^{*}$ improves overall static-ball success over Single-stage but achieves lower success on rolling balls. PAiD achieves the highest overall success under both conditions and retains its advantage at longer distances, although long-range shots remain challenging. These results support PAiD's robustness to variations in ball position and motion when integrated with locomotion.


\begin{figure}[!htbp]
    \vspace{-6pt}
    \centering
    \setlength{\abovecaptionskip}{3pt}
    \includegraphics[width=\linewidth]{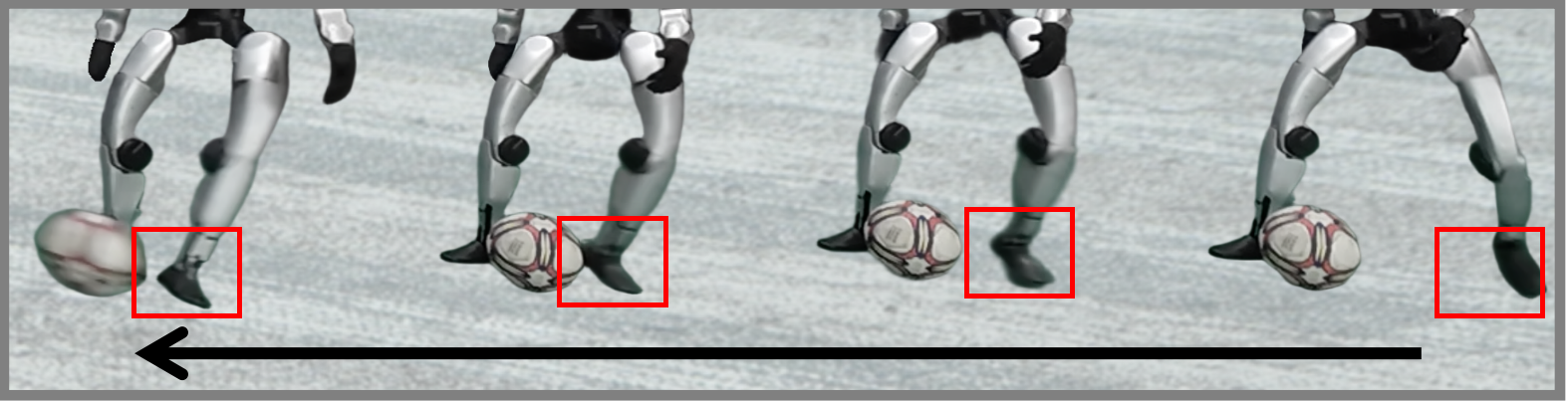}

    \caption{Detailed foot motions during kicking.}
    \label{fig:foot_detail}
    \vspace{-8pt}
\end{figure}

\subsection{Real-World Performance}

To answer \textbf{Q3}, we deploy our shooting policy on the Unitree G1 to evaluate the real-world performance.

\paragraph{Deployment}
We implement a perception-and-decision framework for real robot deployment.
In the perception module, a depth camera D435i is employed to continuously obtain the ball's relative position in the robot frame using YOLOv8 to estimate the ball's mask \cite{yaseen2024yolov8indepthexplorationinternal} when the ball is within the camera’s field of view. To address out-of-view scenarios, Fast-LIO \cite{9372856} with a LiDAR sensor is utilized to estimate the robot's global pose, enabling transformation of the ball's relative position to the world frame. When the ball leaves the camera view, its previous world-frame position is used to recover the relative ball position. The goal center is marked in the map frame and transformed into the robot frame. This integrated RGB-D and LiDAR pipeline provides robust ball localization throughout the shooting process.
In the decision module, we adopt a simple yet effective motion selection mechanism: Before shooting starts, a set of candidate motions is preloaded. The motion whose final-frame robot root position is closest to the estimated ball position is selected and provided as input to a single policy, which is then executed.

\paragraph{Soccer Shooting Proficiency and Terrain Robustness Test}

In Fig.~\ref{fig:big_real_picture} (a), (c), and (d), we set up a goal with a width of 2~m and a height of 1.5~m in the real world. During the experiments, the robot is positioned 5~m in front of the goal, and a soccer ball is randomly placed within an area in front of the robot that roughly matches the depth camera’s field of view, across multiple trials.
Fig.~\ref{fig:real_exp_1} shows that our method achieves a high success rate for most ball locations on both hard ground and grass surfaces, except for a few corner cases where the success rate is lower, as these positions require the robot to perform large body rotations or travel a longer distance.
During testing, the robot successfully kicked soccer balls placed at random positions for up to 11 consecutive trials, demonstrating the robustness of the proposed policy. As shown in Fig.~\ref{fig:foot_detail}, at ball–foot impact, the foot achieves a well-formed strike through the foot arch, which underlies the high shooting success rate.
 In particular, Fig.~\ref{fig:real_exp_1} (c) and (d) demonstrate that our method enables the real robot to successfully kick rolling soccer balls, which is a critical capability for practical soccer gameplay.

We further ablate the two physics-aware transfer components on the real robot. We evaluate each method at ball positions $(-0.6,0.8)$, $(0.0,0.8)$, and $(0.6,0.8)$\,m on hard ground and grass, using 15 trials per position per surface. In Table~\ref{tab:DR_exp}, w/o SystemID disables contact dynamics identification, while w/o Obs Noise disables physics-guided observation noise during training. The Hard Ground and Grass Ground columns report successes/total, while $E_{succ}$ is the percentage of successful trials pooled across both surfaces. Removing either contact dynamics identification or observation noise degrades deployment performance, while combining both yields the strongest transfer across hard-ground and grass settings.

These results demonstrate the high fidelity, expressiveness, and scalability of our method in transferring diverse motions to the real robot.

\section{Conclusion}
\label{sec:conclusion}

In this work, we presented our Perception-Action integrated Decision-making (PAiD) framework for learning robust humanoid soccer shooting skills. By decoupling motion-skill acquisition, perception-action integration, and sim-to-real transfer, our approach overcomes the limitations of single-stage RL and imitation learning. Extensive experiments demonstrate that PAiD achieves superior shooting proficiency and real-world robustness compared to state-of-the-art baselines.

While PAiD shows strong generalization and transferability, its performance at extreme ball positions and under highly dynamic conditions can be further improved. Future work will explore multi-modal perception and adaptive policy architectures to enhance robustness in more complex environments.

\FloatBarrier

\nobalance
\IEEEtriggeratref{52}
\bibliographystyle{IEEEtran}
\bibliography{references}

\end{document}